\documentclass[english]{article}
\usepackage[T1]{fontenc}
\usepackage[latin9]{inputenc}
\usepackage{babel}
\usepackage{url}
\usepackage{amsmath}
\usepackage{graphicx}
\usepackage[unicode=true,pdfusetitle,
 bookmarks=true,bookmarksnumbered=false,bookmarksopen=false,
 breaklinks=false,pdfborder={0 0 1},backref=false,colorlinks=false]
 {hyperref}
\usepackage{breakurl}

\makeatletter

\providecommand{\tabularnewline}{\\}

\usepackage{nips15submit_e,times}
\usepackage{hyperref}
\usepackage{url}
\usepackage{caption}
\usepackage{subcaption}
\usepackage{amsmath}
\usepackage{amsfonts} 
\usepackage{amssymb} 
\usepackage{graphicx}
\usepackage{bm} 
\usepackage{booktabs} 
\usepackage{babel}
\usepackage{comment}

\nipsfinalcopy

\makeatother

\usepackage{listings}

\begin{document}

\title{Optimizing Performance of Recurrent Neural Networks on GPUs}

\author{Jeremy Appleyard$^\star$ \quad Tom\'a\v{s} Ko\v{c}isk\'y$^{\dag\ddag}$ \quad Phil Blunsom$^{\dag\ddag}$ \\ $^\star$NVIDIA \quad $^\dag$University of Oxford \quad $^\ddag$Google DeepMind \\ \texttt{jappleyard@nvidia.com} \\ \texttt{ \{tomas.kocisky,phil.blunsom\}@cs.ox.ac.uk} \\}
\maketitle
\begin{abstract}
As recurrent neural networks become larger and deeper, training times
for single networks are rising into weeks or even months. As such
there is a significant incentive to improve the performance and scalability
of these networks. While GPUs have become the hardware of choice for
training and deploying recurrent models, the implementations employed
often make use of only basic optimizations for these architectures.
In this article we demonstrate that by exposing parallelism between
operations within the network, an order of magnitude speedup across
a range of network sizes can be achieved over a naive implementation.
We describe three stages of optimization that have been incorporated
into the fifth release of NVIDIA's cuDNN: firstly optimizing a single
cell, secondly a single layer, and thirdly the entire network.
\end{abstract}

\section{Introduction}

Recurrent neural networks have become a standard tool for modelling sequential dependencies in discrete time series and have underpinned many recent advances in deep learning, from generative models of images \cite{Gregor:2015:DRAW, Kalchbrenner:2016:GridLSTM} to natural language processing \cite{Kalchbrenner:2013:RCMT,Bahdanau:2014:NMT,Sutskever:2014:SSLNN,Vinyals:2015:ShowTell,Xu:2015:ShowAttendTell,Mikolov:2010:RNNLM,Dyer:2015:StackLSTM}. A key factor in these recent successes has been the availability of powerful Graphics Processing Units (GPUs) which are particularly effective for accelerating the large matrix products at the heart of recurrent networks. However as recurrent networks become deeper \cite{Graves:2012:SSLRNN} and their core units more structured \cite{Hochreiter:1997:LSTM,Cho:2014:GRU}, it has become increasingly difficult to maximally utilise the computational capacity of the latest generation of GPUs.

There have been several studies optimizing the implementation of neural
networks for GPUs, particularly in the case of convolutional neural
networks \cite{DBLP:journals/corr/Lavin15b,DBLP:journals/corr/VasilacheJMCPL14}.
While GPUs are already widely used to compute RNNs \cite{DBLP:journals/corr/AmodeiABCCCCCCD15,DBLP:journals/corr/JozefowiczVSSW16,DBLP:journals/corr/LeonardWW15,JMLR:v16:weninger15a},
there has been less work on the optimization of RNN runtime.

In this article we present a number of options for going beyond straightforward
RNN GPU implementations that allow us to achieve close to maximum
computational throughput for common network architectures. These enhancements
are implemented in the fifth version of NVIDIA's cuDNN library for
Simple RNN, GRU, and LSTM architectures.

\section{Implementation}

In this section we will consider the performance of a forward and
backward propagation passes through an LSTM network\cite{Hochreiter:1997:LSTM}.
This is a standard four-gate LSTM network without peephole connections.
The equations to compute the output at timestep $t$ in the forward
pass of this LSTM are given below:

\begin{flalign*}
\begin{aligned} & i_{i}=\sigma(W_{i}x_{i}+R_{i}h_{t-1}+b_{i})\\
 & f_{t}=\sigma(W_{f}x_{t}+R_{f}h_{t-1}+b_{f})\\
 & o_{t}=\sigma(W_{o}x_{t}+R_{o}h_{t-1}+b_{o})\\
 & c'_{t}=\tanh(W_{c}x_{t}+R_{c}h_{t-1}+b_{c})\\
 & c_{t}=f_{t}\circ c'_{t-1}+i_{t}\circ c'_{t}\\
 & h_{t}=o_{t}\circ\tanh(c_{t})
\end{aligned}
\end{flalign*}

Most of the same strategies found to be beneficial to LSTM performance
are easily transferable to other types of RNN.

\subsection{Naive implementation}

There are many ways to naively implement a single propagation step
of a recurrent neural network. As a starting point we will consider
an implementation where each individual kernel (ie. matrix multiplication,
sigmoid, point-wise addition, etc.) is implemented as a separate kernel.
While the GPU executes the operations within each kernel in parallel,
the kernels are executed back-to-back sequentially. The forward pass
performance of this implementation is poor, achieving approximately
0.4 TFLOPS on a test case with hidden state size 512 and minibatch
64, less than 10\% of the peak performance of the hardware (approximately
5.8 TFLOPS when running at base clocks).\footnote{All runtime and FLOP measurements reported are based off the mean
of 100 executions on an NVIDIA M40 GPU (\url{https://images.nvidia.com/content/tesla/pdf/nvidia-teslam40-datasheet.pdf}),
at default application clocks with auto-boost disabled. The host CPU
is an Intel Xeon CPU E5-2690 v2 @ 3.00GHz.}

A widely used optimization is to combine matrix operations sharing
the same input into a single larger matrix operation. In the forward
pass the standard formulation of an LSTM leads to eight matrix matrix
multiplications: four operating on the recurrent input ($R_{*}h_{t-1}$),
four operating on the input from the previous layer ($W_{*}x_{t}$).
In these groups of four the input is shared, although the weights
are not. As such, it is possible to reformulate a group of four matrix
multiplications into a single matrix multiplication of four times
the size. As larger matrix operations are more parallel (and hence
more efficient), this roughly doubles forward pass throughput to 0.8
TFLOPs in the test case described above. This reformulation is very
easy to implement in most deep learning frameworks leading to its
wide use. A similar optimization is possible for GRU units, with two
groups of three matrices able to be grouped. Single gate RNNs cannot
benefit from this optimization as they have only one matrix multiplication
at each input. The backward pass also benefits from this optimization
as four inputs are transformed into one output. Pseudocode for this
implementation is given in Listing \ref{lis:Pseudocode-demonstrating-forward}.

\begin{lstlisting}[caption={Pseudocode demonstrating the starting point for optimization of the
forward pass.},label={lis:Pseudocode-demonstrating-forward},float,basicstyle={\small\ttfamily},tabsize=2,captionpos=b,frame=lines,boxpos=b]
for l in layers
  for j in iterations
    i(l,j)' = A_i(l) * i(l,j)
    h(l,j)' = A_h(l) * h(l,j)
    for pointwiseOp in pointwiseOps
        do pointwiseOp
\end{lstlisting}

While some of the optimizations that follow have been implemented
before, those that have are by no means universal nor standard practice.

\subsection{Single Cell}

\subsubsection{Streamed matrix operations}

The matrix multiplications performed by RNNs often have insufficient
parallelism for optimal performance on the GPU. Current state-of-the-art
GEMM kernels are implemented with each CUDA block computing a rectangular
tile of the output. The dimensions of these tiles is typically in
the range 32 to 128. Partitioning the matrix multiplications required
for the forward pass of a hidden state size 512, minibatch 64 LSTM
with 128x64 tiles gives a total of 16 CUDA blocks. As blocks reside
on a single GPU streaming multiprocessor (SM), and modern top-of-the-range
GPUs (eg. our M40) currently have 24 streaming multiprocessors, this
matrix multiplication will use at most two thirds of the available
GPU performance. As it is desirable to have multiple blocks per SM
to maximise latency hiding it is clear that to achieve better performance,
it is required that we increase parallelism.

One easy way to increase the parallelism of a single RNN cell is to
execute both matrix multiplications on the GPU concurrently. By using
CUDA streams we can inform hardware that the matrix multiplications
are independent. This doubles the amount of parallelism available
to the GPU, increasing performance by up to 2x for small matrix multiplications.
For larger matrix multiplications, streaming is still useful as it
helps to minimize the so called ``tail effect''. If the number of
blocks launched to the GPU are only sufficient to fill its SMs a few
times they can be thought of as passing through the GPU in waves.
All of the blocks in the first wave finishes at approximately the
same time, and as they finish the second wave begins. This continues
until there is no more work. If the number of waves is small, the
last wave will often have less work to do than the others, creating
a ``tail'' of low performance. By increasing parallelism this tail
can be overlapped with another operation, reducing the performance
penalty.

\subsubsection{Fusion of point-wise operations}

Although parallelism comes naturally to point-wise operations, it
was found that they were being executed inefficiently. This is for
two reasons: firstly because there is a cost associated with launching
a kernel to the GPU; secondly because it is inefficient to move the
output of one point-wise operation all the way out to GPU main memory
before reading it in again moments later for the next. By their nature
point-wise operations are independent, and as such, it is possible
to fuse all of the point-wise kernels into one larger kernel.

\begin{lstlisting}[caption={Pseudocode demonstrating forward pass single cell optimizations.},label={lis:Pseudocode-demonstrating-forward-1},float,basicstyle={\small\ttfamily},tabsize=2,captionpos=b,frame=lines]
for l in layers
  for j in iterations
  set stream 0
  i(l,j)' = A_i(l) * i(l,j)
  set stream 1
  h(l,j)' = A_h(l) * h(l,j)
  wait for stream 0
  do pointwise ops
\end{lstlisting}

\subsection{Single Layer\label{sub:Single-Layer}}

A single recurrent layer comprises many cells, the recurrent input
of each depending on the output of the previous. The input from the
previous layer may not have such a dependency and it is often possible
to concatenate the inputs for multiple time steps producing a larger,
more efficient, matrix multiplication. Selecting the amount of steps
to concatenate over is not trivial: more steps leads to a more efficient
matrix multiplication, but fewer steps reduces the time a recurrent
operation may potentially be waiting for. The exact amount of steps
will depend not only on the hyper-parameters, but also on the target
hardware.

Another operation that is possible, when considering a layer in its
entirety, is re-ordering the layout of the weight matrices. As the
same weight matrices are used repeatedly over the course of a layer
the cost of reordering is typically small compared to the cost of
operating on the matrices. In our tests it was found that pre-transposing
the weight matrix lead to noticeable performance improvements. Note
that because the transpose of the weight matrix is used in the backward
pass this pre-transposition must be performed every pass through the
network.

\begin{lstlisting}[caption={Pseudocode demonstrating optimizations across the forward pass of
a layer.},label={lis:Pseudocode-demonstrating-optimiz},float,basicstyle={\small\ttfamily},tabsize=2,captionpos=b,frame=lines]
for l in layers
  A_it(l) = transpose(A_i(l))
  A_ht(l) = transpose(A_h(l))
  for j in iterations, step s
    set stream 0
    i(l,j:j+s)' = A_it(l) * i(l,j:j+s)
    for k in 1,s
      set stream 1
      h(l,j+k)' = A_ht(l) * h(l,j+k)
      wait for stream 0 to complete operation on j+k
      do pointwise ops
\end{lstlisting}

\subsection{Multiple Layers}

It is becoming increasingly common for RNNs to feature multiple recurrent
layers ``stacked'' such that each recurrent cell feeds its output
directly into a recurrent cell in the next layer. In this situation,
it is possible to exploit the parallelism between recurrent layers:
the completion of a recurrent cell not only resolves the dependency
on the next iteration of the current layer, but also on the current
iteration of the next layer. This allows multiple layers to be computed
in parallel, greatly increasing the amount of work the GPU has at
any given time.

\subsubsection{Scheduling }

As launching work to the GPU takes a small, but not insignificant,
amount of time it is important to consider the order in which the
kernels are launched to the GPU. For example, if GPU resources are
available it is almost always preferable to launch a kernel with all
of its dependencies resolved, rather than a kernel which may be waiting
some time for its dependencies to be cleared. In this way as much
parallelism as possible can be exposed. In order to do this we chose
a simple scheduling rule whereby the next work to be scheduled is
that with the fewest edges to traverse before reaching the ``first''
recurrent cell. If one considers a recurrent network as a 2D grid
of cells, this leads to a diagonal ``wave'' of launches propagating
from the first cell.

\begin{lstlisting}[caption={Pseudocode demonstrating the final optimized forward pass.},label={lis:Pseudocode-demonstrating-the},float,basicstyle={\small\ttfamily},tabsize=2,captionpos=b,frame=lines]
for l in layers
  A_it(l) = transpose(A_i(l))
  A_ht(l) = transpose(A_h(l))

while not Complete
  l,it = get next task
  for j in it->it+s
    set stream 0+2*l
    i(l,j:j+s)' = A_it(l) * i(l,j:j+s)
    for k in 1,s
      set stream 1+2*l
      h(l,j+k)' = A_ht(l) * h(l,j+k)
      wait for stream 0+2*l to complete operation on j+k
      do pointwise ops
\end{lstlisting}

\subsection{Performance}

The impact of each of the optimizations described on the forward pass
of a 1000 step, four layer LSTM network with hidden state size 512
and an input minibatch of 64 is shown in Table \ref{tab:LSTM-forward-pass}.
For this network we achieve an \textasciitilde{}11x speedup over a
completely naive implementation, and a \textasciitilde{}6x speedup
over an implementation with the standard GEMM grouping optimization.

\begin{table}
\centering{}%
\begin{tabular}{|c|c|c|}
\hline 
Optimization & Time per cell (us) & Speedup\tabularnewline
\hline 
\hline 
Naive & 777 & (1.0x)\tabularnewline
\hline 
\#1 Grouped GEMMs & 400 & 1.9x\tabularnewline
\hline 
\#2 Streamed GEMMs & 280 & 2.8x\tabularnewline
\hline 
\#3 Fused point-wise & 146 & 5.3x\tabularnewline
\hline 
\#4 Pre-transpose & 125 & 6.2x\tabularnewline
\hline 
\#5 Batching inputs (2-way) & 119 & 6.5x\tabularnewline
\hline 
\#6 Overlapping layers & 70 & 11.1x\tabularnewline
\hline 
\end{tabular}\protect\caption{LSTM forward pass performance. Each optimization was applied on top
of the previous. These measurements were made on a 100 iteration,
four layer LSTM with a hidden state size of 512 and a minibatch of
64 using cuBLAS 7.5. \label{tab:LSTM-forward-pass}}
\end{table}

\begin{figure}
\includegraphics[scale=1.01]{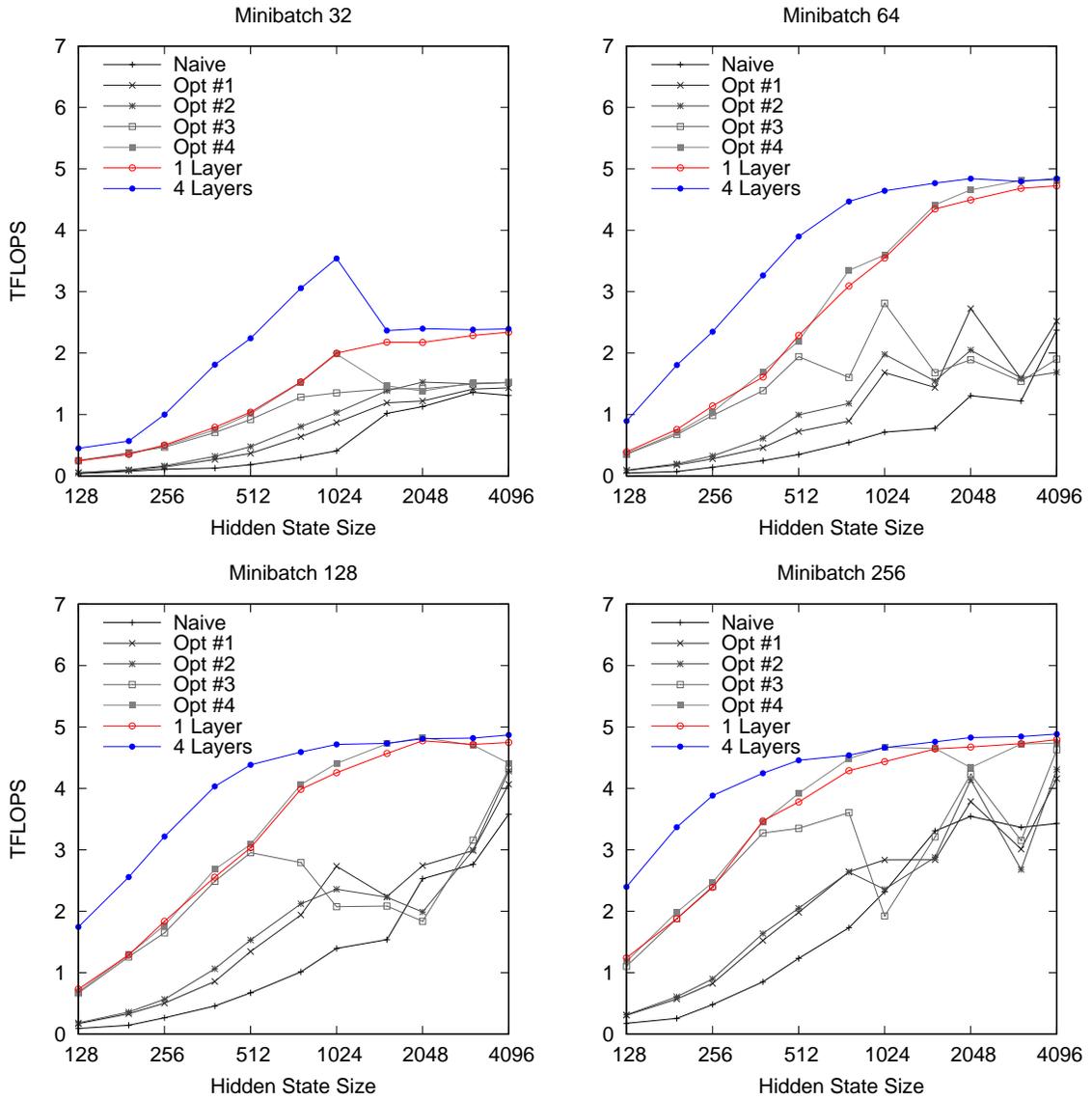}\protect\caption{Impact of optimizations on a the forward pass of a four layer LSTM
network. The peak performance of the M40 GPU used at fixed base clocks
is approximately 5.8 TFLOPS.\label{fig:Impact-of-optimisations}}
\end{figure}

Given sufficient recurrent steps there are three variables that are
expected to significantly influence performance of an RNN implementation.
These are: hidden state size, minibatch size and number of layers.
Fixing the number of layers to four, Figure \ref{fig:Impact-of-optimisations}
shows the impact of each of these optimizations across a wide range
of hidden state sizes and minibatch sizes.

In some cases increasing the number of layers from one to four doubles
throughput (ie. 4x the work in only 2x the time). This performance
improvement is particularly high in low parallelism cases, where the
minibatch is small, the hidden state size is small, or both are small.
One feature of note is the reduction of performance in the minibatch
32 case at high hidden state size. This is attributable to cuBLAS
choosing a different path of execution for matrix multiplications
due to an internal heuristic. As cuBLAS is unaware of the algorithm
it is being used for, this is arguably reasonable behaviour, and performance
for a single layer continues to climb as the problem size increases.

The most significant speedup is seen when the minibatch is 64. As
the minibatch size increases so does the amount of parallelism already
available to the GPU, so optimization strategies focused around increasing
parallelism are less effective. Speedup is also lower with larger
hidden layer sizes, for the same reason. Despite this, even for the
largest problem benchmarked (minibatch 256, hidden state size 4096)
the increase in parallelism due to layer overlapping still brings
better performance.

At larger minibatches it becomes less clear that some of the individual
optimizations bring improvement. Batching inputs is actually found
to be detrimental in many cases for minibatch sizes other than 32,
likely due to the trade-off discussed in Section \ref{sub:Single-Layer}.
In other cases, changes can bring a significant improvement for particular
problem sizes, while causing a slowdown for others. This makes it
hard to say for sure that the combination of all optimizations will
give the fastest runtime for a given set of hyperparameters, however
it is the case that, excepting batched inputs, each optimization helps
in most cases.

\subsection{Weight Update}

The above optimizations only apply to propagation steps. By completing
the gradient propagation before starting the weight update, the weight
update becomes very efficient. A single large matrix multiplication
can be used to update each matrix with no dependencies and this will
usually achieve close to peak performance. Updating the bias weights
is very cheap in comparison to updating the matrices.

\section{cuDNN}

\begin{figure}
\includegraphics{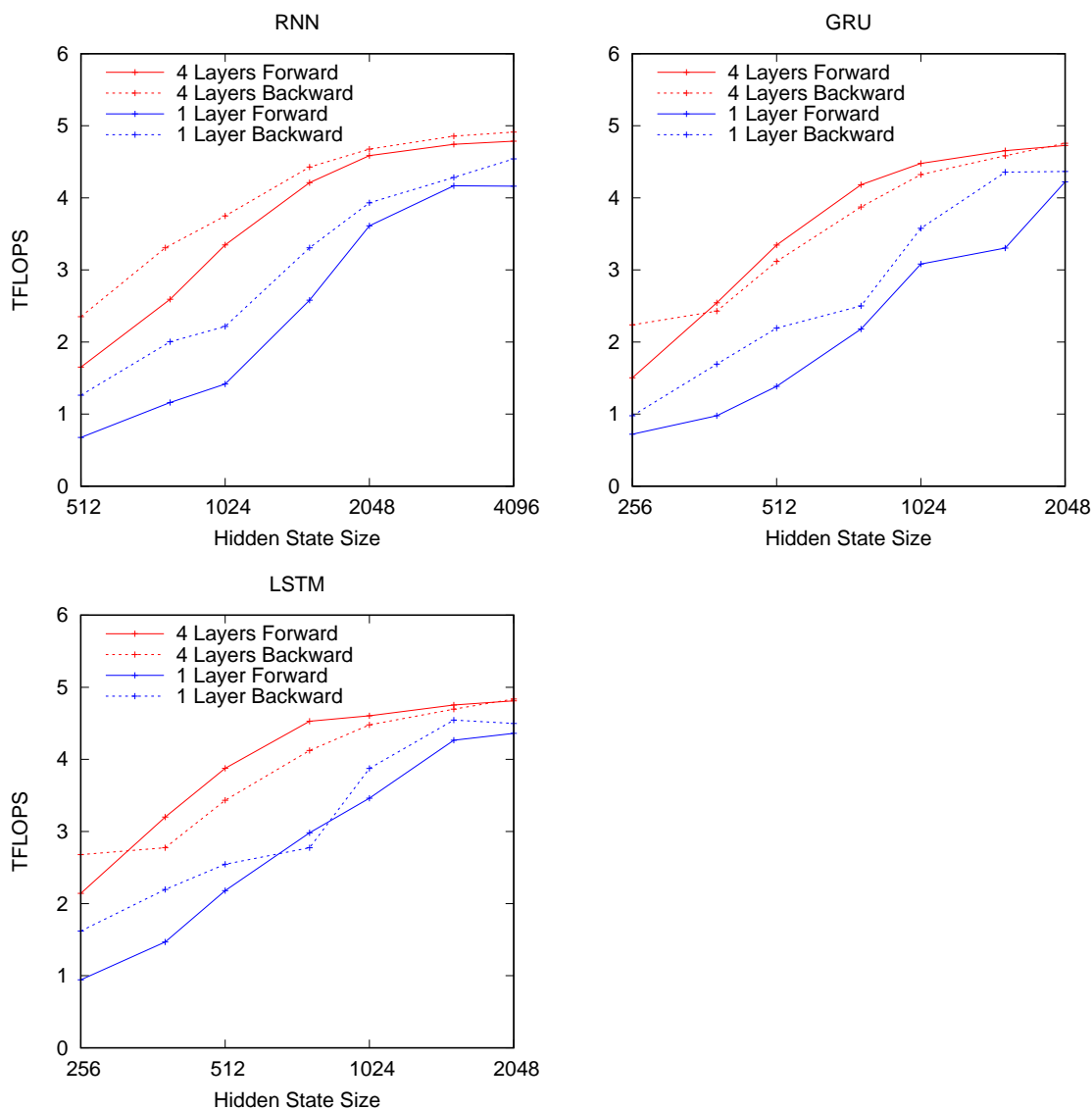}\protect\caption{Forward and backward performance of different networks types using
cuDNN v5 RC. The peak performance of the M40 GPU used at fixed base
clocks is approximately 5.8 TFLOPS.\label{fig:Impact-of-optimisations-1}}
\end{figure}

The optimizations described in Section 2 have been implemented in
the fifth version of NVIDIA's cuDNN library for single gate RNNs,
GRUs and LSTMs. The performance of this implementation is shown in
Figure \ref{fig:Impact-of-optimisations-1}. For this implementation
it was possible to interact at a lower level with cuBLAS than is available
via the current interface, and to tune the heuristics used to determine
the mode of operation of cuBLAS to this use-case. In particular, cuBLAS
will often pick a more parallel but less resource efficient route
if it detects that the GPU is likely to be underused by a call. As
we know the expected amount of parallelism at a higher level than
cuBLAS, overriding this behaviour to favour more resource efficient
paths in cases of high streamed parallelism sometimes resulted in
an overall speedup. It is hoped that an interface to allow this sort
of manual tuning will be incorporated into a future release of cuBLAS.

\section{Conclusions}

We have presented a method by which recurrent neural networks can
be executed on GPUs at high efficiently. While previous implementations
exist achieving good acceleration on GPUs \cite{oxnn,torch-rnn},
to our knowledge none achieve the levels of performance we achieve
using the methods discussed above. The primary strategy used was to
expose as much parallelism to the GPU as possible so as to maximize
the usage of hardware resources. The methods are particularly efficient
when working on smaller deeper recurrent networks where individual
layers have less inherent parallelism.

One feature of the problem that we do not exploit for performance
benefit is the reuse of the parameters between recurrent iterations.
It is conceivable that these parameters could be stored in a lower
level of GPU memory and reused from iteration to iteration. In bandwidth
bound regimes this could potentially greatly improve performance.
There are several drawbacks to this method; firstly the amount of
storage for parameters is limited, and hence there would be an upper
limit on the number of parameters. Secondly, any implementation would
have to make assumptions which are invalid in the CUDA programming
model, and hence would be prone to unexpected failure.

Source code able to reproduce the forward pass timings for each optimization
step is available at \url{https://github.com/parallel-forall/code-samples/blob/master/posts/rnn/LSTM.cu}.
This code closely mirrors the code used to write the core RNN functionality
from version 5 of NVIDIA's cuDNN library, which is available for download
at \url{https://developer.nvidia.com/cudnn}. 

\bibliographystyle{unsrt}
\bibliography{RNNBibliography}

\begin{thebibliography}{10}

\bibitem{Gregor:2015:DRAW}
Karol Gregor, Ivo Danihelka, Alex Graves, Danilo~Jimenez Rezende, and Daan
  Wierstra.
\newblock {DRAW:} {A} recurrent neural network for image generation.
\newblock In {\em Proceedings of the 32nd International Conference on Machine
  Learning, {ICML} 2015, Lille, France, 6-11 July 2015}, pages 1462--1471,
  2015.

\bibitem{Kalchbrenner:2016:GridLSTM}
Ivo~Danihelka Nal~Kalchbrenner, Alex~Graves.
\newblock Grid long short-term memory.
\newblock In {\em Proceedings of the International Conference on Learning
  Representations, {ICLR}, Puerto Rico}, 2016.

\bibitem{Kalchbrenner:2013:RCMT}
Nal Kalchbrenner and Phil Blunsom.
\newblock Recurrent continuous translation models.
\newblock In {\em Proceedings of the 2013 Conference on Empirical Methods in
  Natural Language Processing}, pages 1700--1709, Seattle, Washington, USA,
  October 2013. Association for Computational Linguistics.

\bibitem{Bahdanau:2014:NMT}
Dzmitry Bahdanau, Kyunghyun Cho, and Yoshua Bengio.
\newblock Neural machine translation by jointly learning to align and
  translate.
\newblock In {\em Proceedings of the International Conference on Learning
  Representations, {ICLR}, San Diego}, 2015.

\bibitem{Sutskever:2014:SSLNN}
Ilya Sutskever, Oriol Vinyals, and Quoc V.~V Le.
\newblock Sequence to sequence learning with neural networks.
\newblock In {\em Advances in Neural Information Processing Systems 27}. 2014.

\bibitem{Vinyals:2015:ShowTell}
Oriol Vinyals, Alexander Toshev, Samy Bengio, and Dumitru Erhan.
\newblock Show and tell: {A} neural image caption generator.
\newblock In {\em {IEEE} Conference on Computer Vision and Pattern Recognition,
  {CVPR} 2015, Boston, MA, USA, June 7-12, 2015}, pages 3156--3164, 2015.

\bibitem{Xu:2015:ShowAttendTell}
Kelvin Xu, Jimmy Ba, Ryan Kiros, Kyunghyun Cho, Aaron~C. Courville, Ruslan
  Salakhutdinov, Richard~S. Zemel, and Yoshua Bengio.
\newblock Show, attend and tell: Neural image caption generation with visual
  attention.
\newblock In {\em Proceedings of the 32nd International Conference on Machine
  Learning, {ICML} 2015, Lille, France, 6-11 July 2015}, pages 2048--2057,
  2015.

\bibitem{Mikolov:2010:RNNLM}
Tomas Mikolov, Martin Karafi{\'{a}}t, Luk{\'{a}}s Burget, Jan Cernock{\'{y}},
  and Sanjeev Khudanpur.
\newblock Recurrent neural network based language model.
\newblock In {\em {INTERSPEECH} 2010, 11th Annual Conference of the
  International Speech Communication Association, Makuhari, Chiba, Japan,
  September 26-30, 2010}, pages 1045--1048, 2010.

\bibitem{Dyer:2015:StackLSTM}
Chris Dyer, Miguel Ballesteros, Wang Ling, Austin Matthews, and Noah~A. Smith.
\newblock Transition-based dependency parsing with stack long short-term
  memory.
\newblock In {\em Proceedings of the 53rd Annual Meeting of the Association for
  Computational Linguistics and the 7th International Joint Conference on
  Natural Language Processing}, pages 334--343, Beijing, China, July 2015.
  Association for Computational Linguistics.

\bibitem{Graves:2012:SSLRNN}
Alex Graves.
\newblock {\em Supervised Sequence Labelling with Recurrent Neural Networks},
  volume 385 of {\em Studies in Computational Intelligence}.
\newblock Springer, 2012.

\bibitem{Hochreiter:1997:LSTM}
Sepp Hochreiter and J\"{u}rgen Schmidhuber.
\newblock Long short-term memory.
\newblock {\em Neural Computation}, 9(8):1735--1780, November 1997.

\bibitem{Cho:2014:GRU}
Kyunghyun Cho, Bart van Merrienboer, Dzmitry Bahdanau, and Yoshua Bengio.
\newblock On the properties of neural machine translation: Encoder--decoder
  approaches.
\newblock In {\em Proceedings of SSST-8, Eighth Workshop on Syntax, Semantics
  and Structure in Statistical Translation}, pages 103--111, Doha, Qatar,
  October 2014. Association for Computational Linguistics.

\bibitem{DBLP:journals/corr/Lavin15b}
Andrew Lavin.
\newblock Fast algorithms for convolutional neural networks.
\newblock {\em CoRR}, abs/1509.09308, 2015.

\bibitem{DBLP:journals/corr/VasilacheJMCPL14}
Nicolas Vasilache, Jeff Johnson, Micha{\"{e}}l Mathieu, Soumith Chintala,
  Serkan Piantino, and Yann LeCun.
\newblock Fast convolutional nets with fbfft: {A} {GPU} performance evaluation.
\newblock {\em CoRR}, abs/1412.7580, 2014.

\bibitem{DBLP:journals/corr/AmodeiABCCCCCCD15}
Dario Amodei, Rishita Anubhai, Eric Battenberg, Carl Case, Jared Casper,
  Bryan~C. Catanzaro, Jingdong Chen, Mike Chrzanowski, Adam Coates, Greg
  Diamos, Erich Elsen, Jesse Engel, Linxi Fan, Christopher Fougner, Tony Han,
  Awni~Y. Hannun, Billy Jun, Patrick LeGresley, Libby Lin, Sharan Narang,
  Andrew~Y. Ng, Sherjil Ozair, Ryan Prenger, Jonathan Raiman, Sanjeev Satheesh,
  David Seetapun, Shubho Sengupta, Yi~Wang, Zhiqian Wang, Chong Wang, Bo~Xiao,
  Dani Yogatama, Jun Zhan, and Zhenyao Zhu.
\newblock Deep speech 2: End-to-end speech recognition in {E}nglish and
  {M}andarin.
\newblock {\em CoRR}, abs/1512.02595, 2015.

\bibitem{DBLP:journals/corr/JozefowiczVSSW16}
Rafal J{\'{o}}zefowicz, Oriol Vinyals, Mike Schuster, Noam Shazeer, and Yonghui
  Wu.
\newblock Exploring the limits of language modeling.
\newblock {\em CoRR}, abs/1602.02410, 2016.

\bibitem{DBLP:journals/corr/LeonardWW15}
Nicholas L{\'{e}}onard, Sagar Waghmare, Yang Wang, and Jin{-}Hwa Kim.
\newblock {RNN} : Recurrent library for torch.
\newblock {\em CoRR}, abs/1511.07889, 2015.

\bibitem{JMLR:v16:weninger15a}
Felix Weninger.
\newblock Introducing currennt: The {M}unich open-source {CUDA} recurrent
  neural network toolkit.
\newblock {\em Journal of Machine Learning Research}, 16:547--551, 2015.

\bibitem{oxnn}
Tomas Kocisky.
\newblock oxnn.
\newblock \url{https://github.com/tkocisky/oxnn}, 2015.

\bibitem{torch-rnn}
Justin Johnson.
\newblock Torch-rnn.
\newblock \url{https://github.com/jcjohnson/torch-rnn}, 2016.

\end{thebibliography}

\end{document}